\def\year{2022}\relax
\documentclass[letterpaper]{article} % DO NOT CHANGE THIS
\usepackage{aaai22}  % DO NOT CHANGE THIS
\usepackage{times}  % DO NOT CHANGE THIS
\usepackage{helvet}  % DO NOT CHANGE THIS
\usepackage{courier}  % DO NOT CHANGE THIS
\usepackage[hyphens]{url}  % DO NOT CHANGE THIS
\usepackage{graphicx} % DO NOT CHANGE THIS
\usepackage{natbib}  % DO NOT CHANGE THIS AND DO NOT ADD ANY OPTIONS TO IT
\usepackage{caption} % DO NOT CHANGE THIS AND DO NOT ADD ANY OPTIONS TO IT
\DeclareCaptionStyle{ruled}{labelfont=normalfont,labelsep=colon,strut=off} % DO NOT CHANGE THIS
\usepackage{algorithm}
\usepackage{algorithmic}
\usepackage{amsmath}
\usepackage{amssymb}
\usepackage{multirow}  

\usepackage[font=small]{subfig}
\usepackage{bm}

\usepackage[table]{xcolor}
\definecolor{mygray}{gray}{.92}

\makeatletter
\newcommand{\thickhline}{%
    \noalign {\ifnum 0=`}\fi \hrule height 1pt
    \futurelet \reserved@a \@xhline
}

\newcommand{\blue}[1]{{\color{blue}{#1}}}

\usepackage{newfloat}
\usepackage{listings}
\floatstyle{ruled}
\newfloat{listing}{tb}{lst}{}
\floatname{listing}{Listing}

\usepackage[switch]{lineno}
\usepackage[pagebackref=true,breaklinks=true,letterpaper=true,colorlinks=true,bookmarks=false]{hyperref}

\title{Exploring Visual Context for Weakly Supervised Person Search}
\author{Yichao Yan$^{1}$\thanks{indicates equal contributions},
Jinpeng Li$^{2} \footnotemark[1]$,
Shengcai Liao$^{2}$,
Jie Qin$^{2}$,
Bingbing Ni$^{1}$,
Xiaokang Yang$^{1}$,
Ling Shao$^{2}$
\\
\noindent
$^{1}$ MoE Key Lab of Artificial Intelligence, AI Institute, Shanghai Jiao Tong University, China\qquad \\ $^{2}$ Inception Institute of Artificial Intelligence (IIAI), UAE \qquad \\
$^{}$ {\tt\small \{yanyichao91, ljpadam\}@gmail.com}}
\usepackage{bibentry}
\begin{document}

\maketitle

\begin{abstract}
Person search has recently emerged as a challenging task that jointly addresses pedestrian detection and person re-identification. Existing approaches follow a fully supervised setting where both bounding box and identity annotations are available. However, annotating identities is labor-intensive, limiting the practicability and scalability of current frameworks. This paper inventively considers weakly supervised person search with only bounding box annotations. We proposed to address this novel task by investigating three levels of context clues (i.e., detection, memory and scene) in unconstrained natural images. The first two are employed to promote local and global discriminative capabilities, while the latter enhances clustering accuracy. Despite its simple design, our CGPS achieves 80.0\% in mAP on CUHK-SYSU, boosting the baseline model by 8.8\%. Surprisingly, it even achieves comparable performance with several supervised person search models. Our code is available at {\small \url{https://github.com/ljpadam/CGPS}}.\end{abstract}

\section{Introduction}
Person search \cite{DBLP:conf/cvpr/ZhengZSCYT17,DBLP:conf/cvpr/XiaoLWLW17} aims to retrieve a query person from unconstrained natural images. It therefore needs to simultaneously address the tasks of pedestrian detection and person re-identification (re-id). %In contrast to the conventional person re-id task, which requires the pedestrians to be manually cropped or works with an offline detector, the joint detection and re-id paradigm of person search displays higher efficiency in real scenarios. 
Supervised person search in particular has been extensively studied in recent years \cite{DBLP:conf/iccv/LiuFJKZQJY17,DBLP:journals/pr/XiaoXTHWF19,DBLP:conf/cvpr/MunjalATG19}. Nevertheless, annotating the identities remains labor-intensive. In contrast, relatively accurate bounding box annotations could be automatically generated from existing pedestrian detectors \cite{DBLP:conf/iccv/BrazilYL17,DBLP:conf/eccv/LiuLHLC18,DBLP:conf/cvpr/LiuLRHY19}. Motivated by this, in this work, we consider person search in the weakly supervised setting, in which there only exist bounding box annotations.

\begin{figure}[t]
\begin{center}
    %\fbox{\rule{0pt}{2in} %\rule{0.9\linewidth}{0pt}}
    \includegraphics[width=\linewidth]{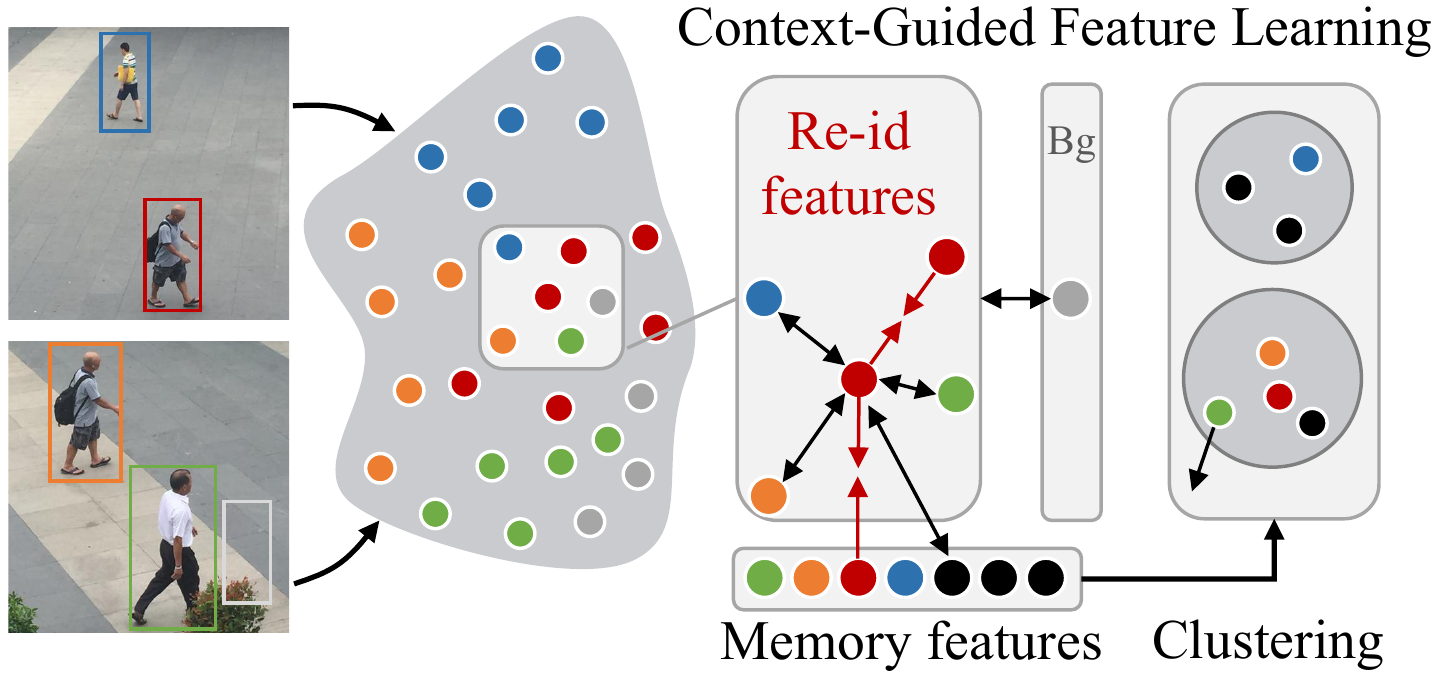}
\end{center}
\vspace{-12pt}
   \caption{Overall pipeline of the proposed context-guided person search framework for weakly supervised person search. %We aim to build a framework for person search with only bounding box annotations, where the re-id embeddings are jointly learned with detection. 
   Without identity annotations, initial pseudo labels (colored points) are generated with ImageNet-pretrained weights. We employ the detection context to pull features belonging to the same identity together, while pushing the re-id features of the pedestrians away from the background features. A hard-negative mining strategy is designed to effectively employ the information in the memory. We use the scene context to generate more accurate clustering results. }
\label{fig:long}
\end{figure}

Intuitively, this task can be addressed with a two-step person search model, by first locating the pedestrians with a detector, and then applying an unsupervised person re-id model. However, there are two major issues with this straightforward solution. \textbf{1)} Two-step approaches employ two separate models to address pedestrian detection and person re-id, requiring twice as many parameters as one-step models, and introducing high computational overhead. Moreover, these models cannot achieve end-to-end inference and are thus inefficient. \textbf{2)} As existing unsupervised re-id methods take cropped person images as input, they do not have access to the rich context information when training the detection model, or the scene context. Therefore, directly extending an existing unsupervised person re-id model to address our task without considering the detection and scene context would make the solution suboptimal.

To address the first issue, in this work, we propose a novel weakly supervised person search framework, termed as the Context-Guided Person Search Network (\textbf{CGPS}). Our model follows the typical architecture of a one-step person search method \cite{DBLP:conf/cvpr/XiaoLWLW17}, but it is only supervised with bounding box annotations. To learn re-id features, we adopt similar spirit of an unsupervised re-id model \cite{DBLP:conf/nips/Ge0C0L20}, and apply the clustering method to acquire human identities for unsupervised re-id training. This framework naturally inherits the advantages of existing one-step person search models in terms of efficiency, while at the same time avoiding the need for labor-intensive identity annotations. 

In terms of the second issue, we further explore the context information for the joint learning framework to pursue discriminative representations for robust unsupervised re-id, as illustrated in Figure~\ref{fig:long}.
Motivated by the recent advances in context learning \cite{DBLP:conf/iccv/DoerschGE15,DBLP:conf/cvpr/0005DSZWTA18}, the following visual context is investigated. \textbf{First}, we explore the \emph{detection context} for re-id feature learning. Detection modules, such as Faster-RCNN \cite{DBLP:journals/pami/RenHG017}, generate multiple positive predictions associated with each ground truth bounding box, as well as the negative predictions corresponding to the background. We observe that these samples can be naturally employed as a local constraint for learning identity features by keeping the re-id features belonging to the same bounding box close, while pushing the person and background features apart. 
\textbf{Second}, we investigate the global \emph{memory context} to enable the model to pay more attention to hard negative samples in the feature memory. We design a sampling strategy based on the model's confidence on the negative samples, which adaptively selects the most effective hard negatives.
\textbf{Third}, we employ the \emph{scene context} to improve the clustering results. Previous methods naively cluster all the re-id features to generate pseudo-labels, without considering the natural context in the scene. We impose an intuitive constraint for clustering, i.e., different people appearing in an image should not belong to the same cluster. We demonstrate that this constraint significantly improves the clustering accuracy. These simple yet effective designs successfully transform an unsupervised person re-id model \cite{DBLP:conf/nips/Ge0C0L20} into a promising weakly supervised person search framework, notably improving the searching performance. 

In summary, our contributions are three-fold: (1) We propose a novel weakly supervised person search framework, which extends the current fully supervised paradigm. In addition, our weakly supervised task can be further extended to a fully unsupervised setting, requiring neither bounding box nor identity annotations on the target dataset.  We expect the pioneering work to encourage future research in this direction. (2) We systematically investigate the visual context clues in the joint learning framework, for both feature learning and clustering. We demonstrate that these context clues work seamlessly in the framework to enhance the re-id feature learning. (3) As a weakly supervised person search framework, our model surprisingly outperforms several fully supervised models on CUHK-SYSU. 
% \begin{itemize}
% % 	\setlength{\itemsep}{0pt}
% % 	\setlength{\parsep}{-2pt}
% % 	\setlength{\parskip}{-0pt}
% % 	\setlength{\leftmargin}{-15pt}
% % 	\vspace{-7pt}
%     \item We propose \textbf{a novel weakly supervised person search framework}, which extends the current fully supervised paradigm. We expect the pioneering work to encourage future research in this direction.  
%     \item We systematically investigate the visual context clues in the joint learning framework, for both feature learning and clustering. We demonstrate that these context clues work seamlessly in the framework to enhance the re-id feature learning.
%     \item As a one-step person search framework, our model outperforms the two-step models under the same supervision on CUHK-SYSU, while displaying remarkably higher efficiency.
%     \item Our weakly supervised task can be further extended to a \textbf{fully unsupervised setting}, requiring neither bounding box nor identity annotations on the target dataset. We believe that the proposed weakly supervised/unsupervised setting complements the conventional supervised one in person search.
% \end{itemize}

\section{Related Work}
\hspace{\parindent} \textbf{Person Search}. Person search is an extension of the person re-id task, which requires pedestrian detection and person re-id to be addressed in uncropped images. \cite{DBLP:conf/mm/XuMHL14} first introduced this task, applying a sliding window search strategy with handcrafted features. However, this task did not draw much attention, due to lack of proper benchmarks, until \cite{DBLP:conf/cvpr/ZhengZSCYT17} introduced a large dataset for person search, and systematically evaluated different combinations of several pedestrian detectors and re-id models. Meanwhile, \cite{DBLP:conf/cvpr/XiaoLWLW17} proposed the first one-step detection and re-id network, enabling end-to-end learning. The follow-up works that have since been introduced can thus be grouped into one-step and two-step models. In general, two-step models \cite{DBLP:conf/eccv/ChenZOYT18,DBLP:conf/eccv/LanZG18,DBLP:conf/iccv/HanYZTZGS19,DBLP:conf/cvpr/WangMCSC20} pay more attention to the consistency issue, i.e., how to learn discriminative re-id features based on the detection results. In contrast, one-step models \cite{DBLP:conf/iccv/LiuFJKZQJY17,DBLP:conf/eccv/ChangHSLYH18,DBLP:conf/cvpr/DongZST20a,DBLP:conf/cvpr/ChenZYS20} focus more on how to design an efficient person search framework. 

Despite their impressive progress, most of the current models are fully supervised. In this work, we explore the weakly supervised setting, i.e., we aim to learn the person search model with only bounding box annotations, while the re-id task is learned in an unsupervised manner.

\textbf{Unsupervised Person Re-id}. Early works, such as ELF \cite{DBLP:conf/eccv/GrayT08}, LBP \cite{DBLP:conf/eccv/XiongGCS14} and LOMO \cite{DBLP:conf/cvpr/LiaoHZL15}, typically resort to handcrafted features. However, these models only achieve limited performance. Deep learning-based methods can generally be divided into two categories: (1) those that generate pseudo-labels for unsupervised instances \cite{DBLP:conf/aaai/LinD00019,DBLP:conf/cvpr/ZengNW020}, and (2) those that translate labeled examples into the unlabeled domain \cite{DBLP:conf/cvpr/SongYSXH19,DBLP:conf/iccv/FuWWZSUH19,DBLP:conf/eccv/MekhazniBEG20}. For the first category, pseudo-labels are typically generated by clustering the re-id embeddings. If videos are available, temporal constraints \cite{DBLP:conf/bmvc/ChenZG18,DBLP:journals/pami/LiZG20} can be imposed to improve the clustering results. The second category is unsupervised domain adaptation, where the models are trained on the source domain and transferred to the unsupervised target domain. This can be achieved with generative adversarial networks \cite{DBLP:conf/cvpr/WeiZ0018,DBLP:conf/cvpr/Deng0YK0J18,DBLP:conf/cvpr/LiuCS20,DBLP:conf/iccv/ChenZG19,DBLP:conf/iccv/HuangWXZ19}, clustering \cite{DBLP:conf/cvpr/Zhong0LL019,DBLP:conf/iccv/FuWWZSUH19,DBLP:conf/cvpr/ZhaiLYSCJ020}, and soft labels assigning \cite{DBLP:conf/cvpr/WangZ20a}. 

\begin{figure*}[t]
\begin{center}
    \includegraphics[width=0.9\linewidth]{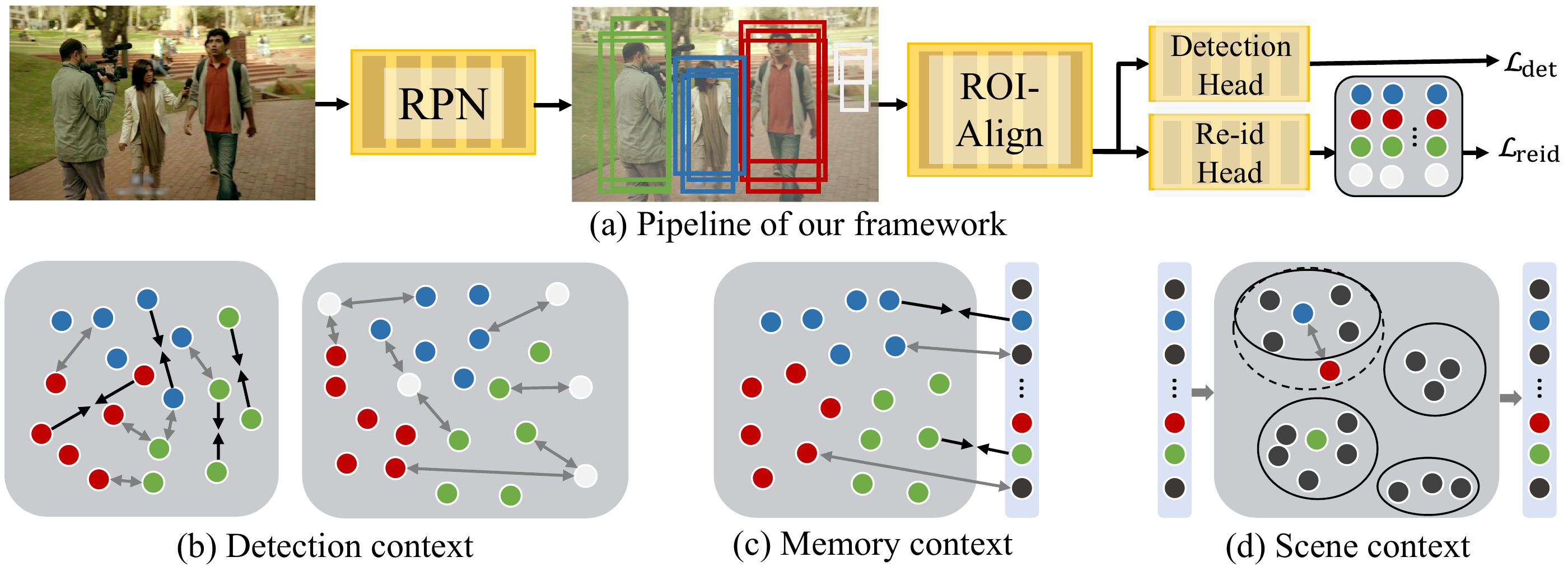}
\end{center}
\vspace{-12pt}
   \caption{Illustration of our framework. (a) The basic architecture of our framework follows the typical person search model \cite{DBLP:conf/cvpr/XiaoLWLW17}, and we employ an unsupervised method \cite{DBLP:conf/nips/Ge0C0L20} to train the re-id branch. (b) We employ the local context information generated by the detection task, to pull positive features belonging to the same instance together, and push background features apart. (c) We sample the hard negatives in the memory to facilitate global feature comparison. (d) Scene context is employed to generate more accurate clustering results. }
\label{fig:method}
\end{figure*}

In this work, we consider the task without any re-id annotations. Therefore, we employ clustering to generate pseudo-labels for the re-id task. Furthermore, we extensively explore the context information in the joint learning framework to improve the quality of pseudo-labels and to enhance the discriminative ability of the re-id features.

\textbf{Learning Visual Context}. In addition to annotations, visual data implicitly contains rich context information, which has been widely explored in various vision tasks. For example, the relative location of object parts can be explicitly employed for visual representation learning \cite{DBLP:conf/iccv/DoerschGE15,DBLP:conf/eccv/NorooziF16}. Meanwhile, the spatial and semantic relations between objects are helpful for the detection \cite{DBLP:conf/cvpr/DivvalaHHEH09,DBLP:conf/eccv/DvornikMS18,DBLP:conf/cvpr/BarneaB19} and segmentation \cite{DBLP:conf/cvpr/MottaghiCLCLFUY14,DBLP:conf/cvpr/0005DSZWTA18} tasks. In videos, the spatial-temporal relations can be modeled with a graph model to encode the interactions between objects, and enhance the performance of action recognition \cite{DBLP:conf/eccv/WangG18,DBLP:conf/cvpr/LiCCZW019}. Some person recognition models \cite{DBLP:conf/cvpr/HuangXL18,DBLP:conf/cvpr/YanZNZXY19} have also recently employed the social/group context, to pursue robust identity representation.

In our work, we simultaneously investigate three levels of context information, i.e., the detection context, the memory context, and the scene context, and further reveal their importance in weakly supervised person search.

\section{Methodology}
In this section, we first describe the overall framework for weakly supervised person search, and then delve into the details of the proposed context-guided learning strategies.
%Finally, we provide the network details.

\subsection{Framework Overview}\label{sec:fo}

\textbf{Person Search Framework}. In this work, we adopt the typical person search framework \cite{DBLP:conf/cvpr/XiaoLWLW17} which jointly learns detection and re-id in an end-to-end manner. Specifically, this framework is developed upon the Faster-RCNN \cite{DBLP:journals/pami/RenHG017} architecture, as shown in Figure \ref{fig:method}(a). Given an input image, the region proposal network (RPN) generates a set of pedestrian proposals, which are subsequently input into an RoI-Align layer to further extract the features of the candidate proposals. The output features are passed to a detection head and a re-id head, respectively. The detection head is trained with a classification loss and a regression loss, which supervise the person/non-person scores, as well as the bounding box locations. Meanwhile, the re-id head outputs the identity embedding associated with each pedestrian proposal, which is supervised with the pseudo-labels generated in an unsupervised manner. 

\textbf{Unsupervised Person Re-id}. Among the recent methods \cite{DBLP:conf/cvpr/LinXWY020,DBLP:conf/aaai/LinD00019,DBLP:conf/cvpr/ZengNW020,DBLP:conf/nips/Ge0C0L20} designed for unsupervised re-id, we employ the memory-based loss \cite{DBLP:conf/nips/Ge0C0L20} in our framework, for two reasons. First, it achieves strong performance on several unsupervised person re-id benchmarks, and is more likely to work well in our task. Second, as has been demonstrated in OIM \cite{DBLP:conf/cvpr/XiaoLWLW17}, memory-based learning schemes can effectively investigate the information in the unlabeled identities for person search. Specifically, the re-id features of all the training instances are stored in a memory $\bm{V} \in \mathbb{R}^{D \times N_a} = \{\bm{v}_1,...,\bm{v}_{N_a}\}$, which contains $N_a$ feature vectors with dimension $D$, where $N_a$ is the number of instances in the training set. %By performing clustering, we will get $n_c$ cluster centroids $\bm{C} \in \mathbb{R}^{D \times n_c} = \{\bm{c}_1,...,\bm{c}_{n_c}\}$, and $n_u$ unclustered instances $\bm{U} \in \mathbb{R}^{D \times n_u} = \{\bm{u}_1,...,\bm{u}_{n_u}\}$. Given the re-id feature $\bm{x}$ for the $i$-th instance, the loss is defined as:
% \vspace{-4pt}
% \begin{equation}\small\label{eq:NCE}
%     \mathcal{L}_i= {\rm -log}\frac{{\rm exp}( \bm{x} \cdot  \bm{v}^{+} / \tau)}{\sum_{j=1}^{n_c}{\rm exp}(\bm{x} \cdot  \bm{c}_j / \tau) + \sum_{k=1}^{n_u}{\rm exp}(\bm{x} \cdot  \bm{u}_k / \tau)},
% \vspace{-3pt}
% \end{equation}
Through clustering, we obtain $N_c$ clusters $\{\mathbb{C}_1, ..., \mathbb{C}_{N_c}\}$ with centroids $\bm{C} \in \mathbb{R}^{D \times N_c} = \{\bm{c}_1,...,\bm{c}_{N_c}\}$. Note that we do not differentiate the clusters containing multiple instances from those with only one (i.e., the unclustered instance). Suppose the network outputs a re-id feature $\bm{x}_i$, the loss is defined as:
% \vspace{-4pt}
\begin{equation}\small\label{eq:NCE}
    \mathcal{L}_i= {\rm -log}\frac{{\rm exp}( \bm{x}_i \cdot  \bm{c}^{+} / \tau)}{\sum_{j=1}^{N_c}{\rm exp}(\bm{x}_i \cdot  \bm{c}_j / \tau) },
% \vspace{-3pt}
\end{equation}
where $\bm{c}^{+} = \bm{c}_j$ if $\bm{x}_i$ belongs to the $j$-th cluster, `$\cdot$' denotes the inner product, and $\tau\!>\!0$ is a temperature hyper-parameter that controls the softness of the probability distribution. During backpropagation, the corresponding memory feature is updated by:
\begin{equation}\small\label{eq:update}
\bm{v} \leftarrow \gamma \bm{v} + (1-\gamma)\bm{x}_i,
% \vspace{-3pt}
\end{equation}
where $\gamma \in [0, 1]$ controls the update ratio in the memory. 

\subsection{Context-Guided Feature Learning}\label{sec:cg}
We observe that the unsupervised re-id model trained with cropped instances does not make full use of the context information in person search. To better adapt the unsupervised re-id to person search, we investigate three types of context information, i.e., the detection context, the memory context, and the scene context.

\textbf{Detection Context}. When training the person search model, the RPN will generate several positive samples for each person, as well as some negative samples corresponding to the background. However, the loss function in Eq. \ref{eq:NCE} only employs the positive features, which yields two issues. 1) The input re-id features are only compared with memory features, while relations with the intra-batch re-id features are not explored. 2) There is no constraint on the background and foreground features. Although a few supervised person search models \cite{DBLP:conf/cvpr/ChenZYS20,DBLP:conf/aaai/ChenZO0S20} have addressed the second issue, the first issue tends to play a more important role in the unsupervised scenario. In this work, we propose a quadruplet loss to simultaneously address both issues. Specifically, suppose the model outputs $n_x$ re-id features $\bm{X} \in \mathbb{R}^{D \times n_x} = \{\bm{x}_1,...,\bm{x}_{n_x}\}$ for the pedestrians, and $n_b$ features corresponding to the backgrounds $\bm{B} \in \mathbb{R}^{D \times n_b} = \{\bm{b}_1,...,\bm{b}_{n_b}\}$. For an instance $\bm{x}_i \in \bm{X}$, the loss function is defined as:
% \vspace{-4pt}
\begin{equation}\small\label{eq:dc}
\begin{aligned}
    \mathcal{L}_i^{\rm DC} = &-\alpha_{1} [{\rm min}(\bm{x}_{i} \cdot \bm{x}^{+}) - {\rm max}(\bm{x}_i \cdot \bm{x}^{-}) -m] \\
    &- \alpha_{2}[{\rm min}(\bm{x}_{i} \cdot \bm{x}) - {\rm max}(\bm{x}_i \cdot \bm{b}) - m],
\end{aligned}
% \vspace{-3pt}
\end{equation}
where $\bm{x}^{+}$ and $\bm{x}^{-}$ denote the positive and negative samples compared with $\bm{x}_i$, $\bm{x} \in \bm{X}$ and $\bm{b} \in \bm{B}$, while $m$ denotes the distance margin, and $\alpha_1$ and $\alpha_2$ are the weights to balance the two loss terms. As illustrated in Figure \ref{fig:method}(b), this function contains two terms. 1) An instance-to-instance term, which pulls the features belonging to the same instance close together, while pushing apart the features of different instances. 2) An instance-to-background term, which pushes the foreground features away from the background ones. In this way, we provide explicit instance-level guidance on the re-id feature, to yield more discriminative representations.

\textbf{Memory Context}. We observe that the negative samples contribute equally in Eq. \ref{eq:NCE}. However, as indicated by prior works \cite{DBLP:conf/cvpr/ZengNW020,DBLP:conf/aaai/ChenZO0S20}, putting more focus on the hard examples plays an important role in improving the model's discriminative capability. In this work, we explore the hard negative features in the memory, and propose a hard negative sampling strategy by evaluating the cumulative confidence of the hard-negative samples.
Without loss of generality, all the negative features are sorted in descending order $\{\bm{c}^{-}_1,...,\bm{c}^{-}_{N^-}\}$ according to their similarities with the input feature, where the number of negative samples is $N^- = N_c - 1$, indicating that the input feature only belongs to one specific cluster. We determine the number of hardest negative samples with: 
% \vspace{-4pt}
\begin{equation}\small
K = {\mathop{\arg\min}_{k}} | \frac{\sum_{z=1}^{k} \bm{x}_i \cdot \bm{c}^{-}_z}{\sum_{j=1}^{N^-}\bm{x}_i \cdot\bm{c}^{-}_j} - \lambda|, 
% \vspace{-3pt}
\end{equation}
where $\lambda$ is a threshold that controls the ratio of hard negative samples. In this case, Eq. \ref{eq:NCE} becomes:
% \vspace{-4pt}
\begin{equation}\small\label{eq:mc}
    \mathcal{L}_i^{\rm MC}= {\rm -log}\frac{{\rm exp}( \bm{x}_i \cdot  \bm{c}^{+} / \tau)}{ {\rm exp}(\bm{x}_i \cdot  \bm{c}^{+} / \tau) + \sum_{k=1}^{K}{\rm exp}(\bm{x}_i \cdot  \bm{c}^{-}_k / \tau)}.
% \vspace{-3pt}
\end{equation}
Compared to sampling a fixed size of hard negative samples, our sampling strategy is more flexible to focus on a small set of the most effective negative samples. 

The overall loss function is a combination of the targets in Eq. \ref{eq:dc} and Eq. \ref{eq:mc}:
% \vspace{-4pt}
\begin{equation}\small
\mathcal{L}_{\rm reid} = \sum_{i}(\mathcal{L}_i^{\rm DC} + \mathcal{L}_i^{\rm MC}).
% \vspace{-3pt}
\end{equation}
We find these two terms are complementary to each other: $\mathcal{L}_i^{\rm DC}$ tries to discriminate the local features within a batch, while $\mathcal{L}_i^{\rm MC}$ focuses more on the global discrimination by comparing the output feature with all the instances in the memory. Therefore, combining them is likely to yield more discriminative representations.

\renewcommand{\algorithmiccomment}[1]{\hfill$\triangleright$\textit{#1}}

\setlength{\textfloatsep}{10pt}
\begin{algorithm}[t]
\caption{Clustering with Scene Context}
\label{alg:1}
\begin{algorithmic}[1]
\REQUIRE  $\bm{I}$, $\bm{V}$ \\
%\ENSURE Hypergraph feature $O_p$ \\
\STATE Cluster $\bm{V} $ with DBSCAN $\rightarrow \mathbb{C}_1, ..., \mathbb{C}_{N_c} $
\FOR[\blue{iterate each cluster}]{$i \leftarrow 1,...,N_c$} 
\FOR[\blue{iterate each image}]{$j \leftarrow 1,...,N_I$}
\STATE  
$\{\bm{v}_{i,j}^{1},...,\bm{v}_{i,j}^{K}\} \leftarrow \mathbb{C}_i \cap I_j$ \\
\algorithmiccomment{\blue{find the instances belonging to $\mathbb{C}_i$ and $I_j$}}
\STATE  
$ l \leftarrow {\mathop{\arg\max}_{k}}( \bm{v}_{i,j}^{k} \cdot \bm{c}_i )$  \\
\algorithmiccomment{\blue{find the instance nearest to the cluster center}}

\FOR{$k \leftarrow 1,...,l-1, l+1,...,K$} 
\STATE $\mathbb{C}_{i} \leftarrow \mathbb{C}_{i} - \bm{v}_{i,j}^{k}$ \algorithmiccomment{\blue{remove this instance from $\mathbb{C}_{i}$}}
\STATE $N_c \leftarrow N_c + 1$ \algorithmiccomment{\blue{update the number of clusters}}
\STATE $\mathbb{C}_{N_c} \leftarrow \{\bm{v}_{i,j}^{k}\}$ \algorithmiccomment{\blue{add a new cluster}}
\ENDFOR
\ENDFOR
\ENDFOR
\end{algorithmic}
\end{algorithm}

\textbf{Scene Context}. To assign pseudo-labels for the training instances, existing works \cite{DBLP:conf/nips/Ge0C0L20,DBLP:conf/eccv/ChenLL020} directly perform clustering (e.g., DBSCAN \cite{DBLP:conf/kdd/EsterKSX96}) on the instance-level re-id features. However, the results of the unsupervised clustering are not perfect, which directly influences the re-id feature learning in the framework. Although some works have tried to improve the clustering reliability with self-paced learning \cite{DBLP:conf/nips/Ge0C0L20} or instance discrimination learning \cite{DBLP:conf/eccv/ChenLL020}, these works only consider regularizing the instance-level representation, neglecting the relative relations between examples in the scene. In this work, we impose a simple constraint on the clustering results: \emph{the persons appearing in the same image cannot belong to the same cluster}. Specifically, after generating the clustering results based on the memory features, we further go through each cluster and split the clusters containing multiple instances belonging to the same image. For a certain image, only a single instance that is closest to the cluster center will be retained, while other instances will be removed and become unclustered instances. As shown in Figure \ref{fig:method} (d), the original top left cluster contains the blue and red instances. However, since they belong to the same image, the red one, which is far away from the center, is excluded from the cluster. In this way, we are able to generate better clustering results without any additional annotation. Suppose we have $N_I$ training images, and $I_i$ contains the instances in the $i$-th image. Give $\bm{I} = \{I_1,...,I_{N_I}\}$ and the memory feature $\bm{V}$, we summarize the clustering with scene context in Algorithm \ref{alg:1}. 

\subsection{Discussions}
Although some similar spirits have been adopted in prior supervised models, our work is essentially different from them. \emph{First}, TCTS~\cite{DBLP:conf/cvpr/WangMCSC20} and AlignPS~\cite{Yan_2021_CVPR} employ the within-image relation in the training loss, while we consider this as a constraint for generating more accurate pseudo labels. \emph{Second}, although several supervised models~\cite{DBLP:conf/aaai/ChenZO0S20,DBLP:conf/cvpr/ChenZYS20} consider pushing the person and background features apart, our quadruplet loss additionally explores the relations of intra-batch re-id features. \emph{Third}, we investigate three context clues to explicitly tackle the task of weakly supervised person search, while existing model focus on fully supervised setting.

% \subsection{Network and Implementation Details}\label{sec:3_3}
% We employ ResNet-50 \cite{DBLP:conf/cvpr/HeZRS16} as our backbone, which is divided into two stages following Faster-RCNN \cite{DBLP:journals/pami/RenHG017}. Specifically, the first stage contains the layers from $\texttt{conv1}$ to $\texttt{conv4}$, where the features are fed into the RPN to generate a number of region proposals. Then, for each proposal, the RoI-Align layer generates a fixed size ($14 \times 14 \times 1024$) feature map, which is subsequently fed into the second part ($\texttt{conv5}$ layers) of ResNet-50. Finally, we perform average pooling to the output feature map, resulting in a 2048-dimensional feature vector for each proposal. This vector is then fed into the detection head and the re-id head, respectively. The detection head contains a binary $\texttt{Softmax}$ layer for person/non-person classification, as well as a four-dimensional fully connected layer for bounding box regression. Meanwhile, the re-id head contains a 256-dimensional fully connected layer, and the features are L2 normalized for re-id feature learning.

\begin{figure*}[t]
\setlength{\abovecaptionskip}{2mm}
\begin{center}
    %\fbox{\rule{0pt}{2in} %\rule{0.9\linewidth}{0pt}}
   \includegraphics[width=0.45\linewidth]{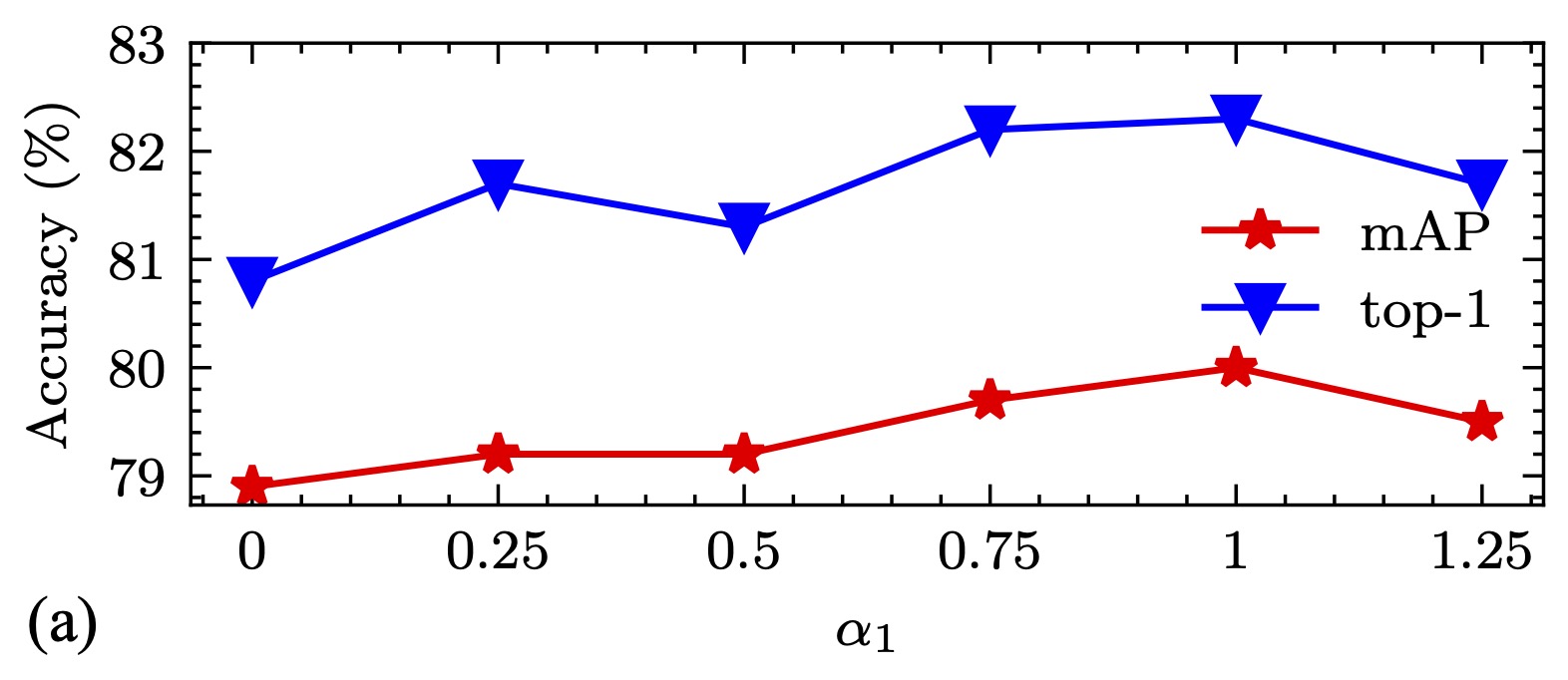}
   \hspace{8mm}
   \includegraphics[width=0.45\linewidth]{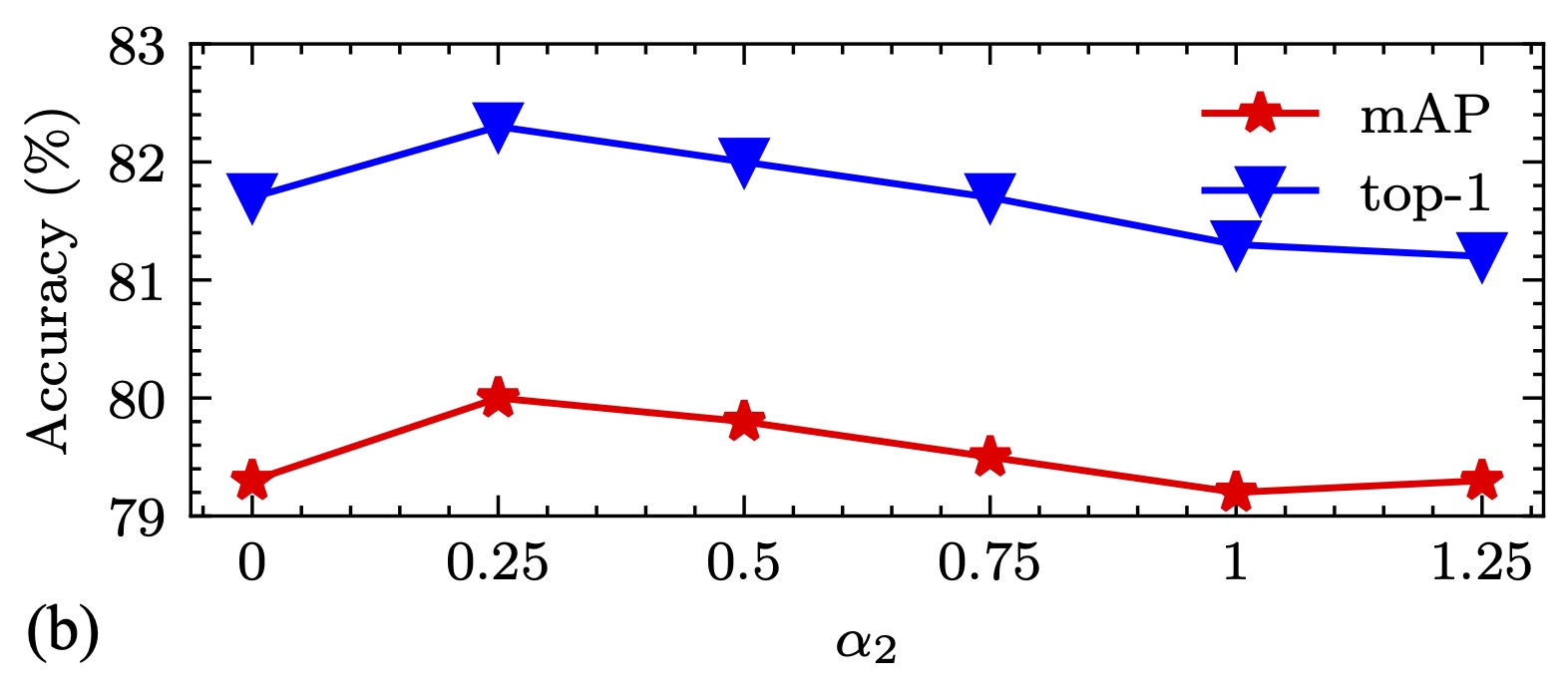}
   \includegraphics[width=0.45\linewidth]{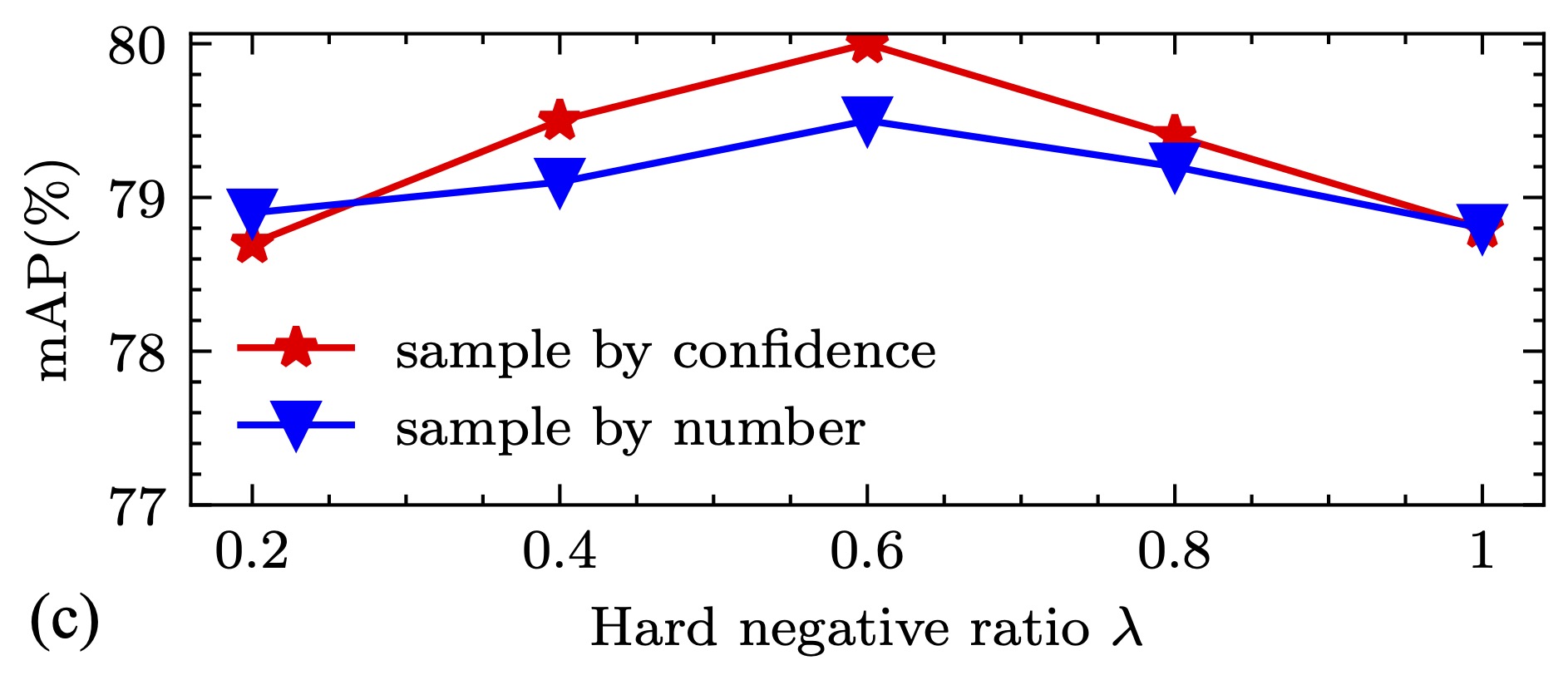}
    \hspace{8mm}
   \includegraphics[width=0.45\linewidth]{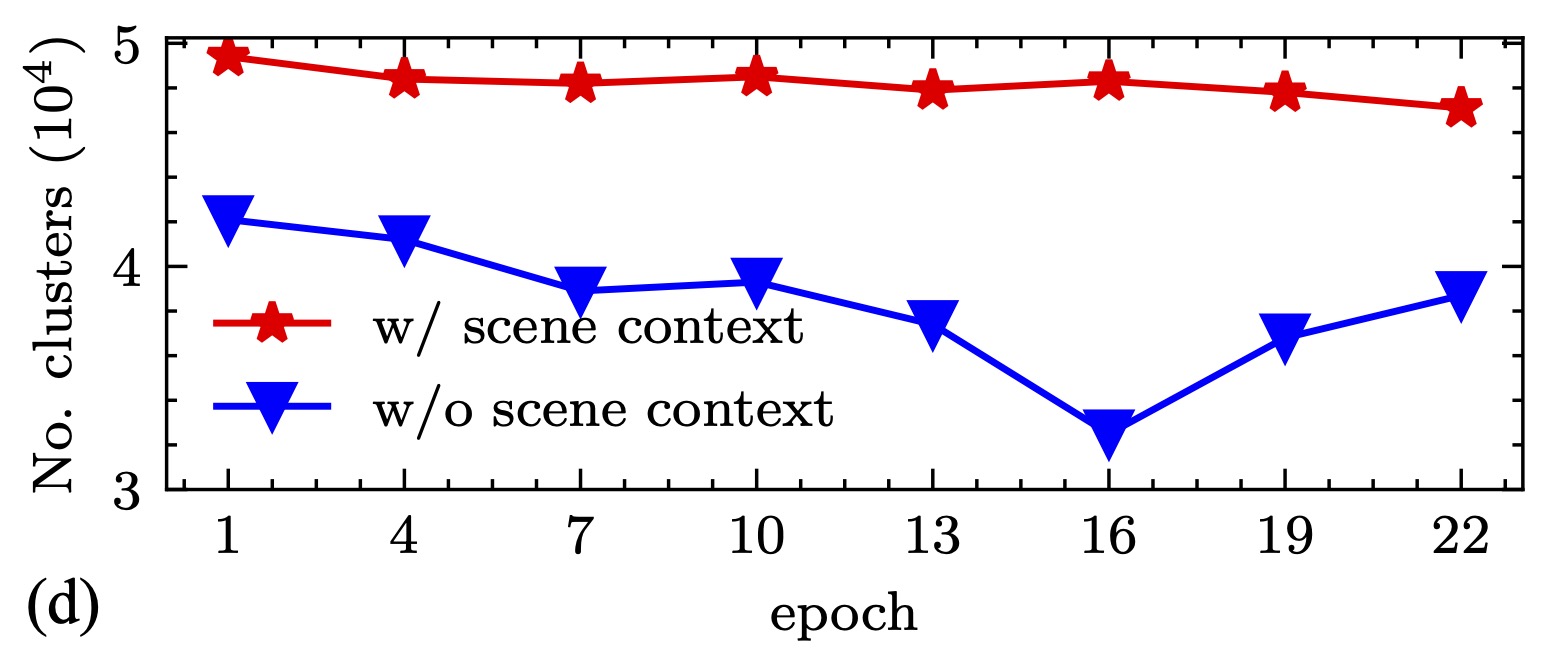}

\end{center}
\vspace{-12pt}
   \caption{Visualization of the contribution of different types of context clues on CUHK-SYSU. (a) The contribution of the instance-to-instance term in the detection context. (b) The impact of the instance-to-background term in the detection context. (c) The influence of hard negative sampling. (d) The contribution of scene context.}
\label{fig:contexts}
\vspace{-4mm}
\end{figure*}

\section{Experiments}
\subsection{Experimental Setup}
\textbf{Datasets}. We evaluate the proposed framework on the following two datasets. \textbf{CUHK-SYSU} \cite{DBLP:conf/cvpr/XiaoLWLW17} is one of the largest public datasets designed for person search, which contains 18,184 images captured from streets and TV/movie frames. This dataset also includes 96,143 bounding box annotations, with 8,432 different identities. \textbf{PRW} \cite{DBLP:conf/cvpr/ZhengZSCYT17} was collected from six surveillance cameras. It contains 11,816 video frames, with 43,110 annotated bounding boxes and 932 identities.  

\noindent\textbf{Evaluation Protocol}. We employ the standard \texttt{train/test} splits for both CUHK-SYSU and PRW. Only the bounding box annotations are employed in the training phase, i.e., 11,216 images with 55,272 bounding boxes for CUHK-SYSU, and 5,705 images with 18,048 boxes for PRW. The test set of CUHK-SYSU contains 2,900 query persons and 6,978 images. In the following sections, we report the results with 100 gallery images unless otherwise specified. PRW contains a test set with 2,057 queries and 6,112 images. We report the mean average precision (mAP) and top-1 ranking accuracy as evaluation metrics.

\noindent\textbf{Implementation Details}
% We initialize the backbone weights with a model pretrained on ImageNet \cite{DBLP:conf/cvpr/DengDSLL009}. 
Following~\cite{Yan_2021_CVPR}, we employ a multi-scale training strategy, while input images are resized to a fixed size of $1500 \times 900$ for inference. In the training phase, random flipping is applied and we employ stochastic gradient descent (SGD) optimizer, with the weight decay set to 0.0005. We set the batch size to 4 and initialize the learning rate to 0.0012, which is reduced by a factor of 10 at epoch 16, training to a total of 22 epochs. We set the default hyperparameters $\gamma = 0.2$, $\alpha_1 = 1$, $\alpha_2 = 0.25$, and $\lambda = 0.6$. We employ DBSCAN \cite{DBLP:conf/kdd/EsterKSX96} with self-paced training \cite{DBLP:conf/nips/Ge0C0L20} as the clustering method. We set $\epsilon=0.7$, while other hyperparameters follow SPCL \cite{DBLP:conf/nips/Ge0C0L20}. All the experiments are implemented in Pytorch, with a Tesla V100 GPU.

\subsection{Analytical Results} \label{sec:ar}
\textbf{Comparative Results}. 
We first evaluate the effectiveness of the proposed context-guided learning strategies. We compare the baseline method with different combinations of context information, and report the results on CUHK-SYSU in Table \ref{tab:cuhk1}. We observe that the baseline framework achieves 71.2\% mAP and 73.8\% top-1 accuracy without any identity annotation. With the help of detection, memory and scene context, the model achieves 4.3\%, 0.1\% and 5.9\% improvements in mAP, respectively. Furthermore, by combining these contexts, the model's performance is further consistently improved. Finally, the model achieves 80.0\% mAP when employing all the context, significantly improving the baseline model by \textbf{8.8}\% in mAP.

\begin{table}[t]
\setlength{\abovecaptionskip}{2mm}
\centering
\begin{tabular}{p{1.2cm}<{\centering}p{1.1cm}<{\centering}p{1.0cm}<{\centering}|p{1.5cm}<{\centering}p{1.5cm}<{\centering}}
\hline\thickhline
\rowcolor{mygray}  
Detection              & Memory               & Scene              & \multicolumn{2}{c}{CUHK-SYSU}                        \\ \cline{4-5} 
\rowcolor{mygray}  
context                 & context                 & context               & mAP                  & \multicolumn{1}{c}{top-1} \\ 
\hline \hline  
 &   & & 71.2 & 73.8                         \\ 
$\surd$ &   & & 75.5$\ $(\blue{+4.3}) & 78.6$\ $(\blue{+4.8})                      \\
 & $\surd$  & & 71.3$\ $(\blue{+0.1}) & 74.1$\ $(\blue{+0.3})                           \\
  &   & $\surd$ & 77.1$\ $(\blue{+5.9}) & 79.9$\ $(\blue{+6.1})                           \\
$\surd$ & $\surd$  & & 75.8$\ $(\blue{+4.6}) & 78.8$\ $(\blue{+5.0})                           \\
$\surd$ &   & $\surd$ & 78.8$\ $(\blue{+7.6}) & 80.9$\ $(\blue{+7.1})                           \\
 & $\surd$  & $\surd$& 78.7$\ $(\blue{+7.5}) & 81.2$\ $(\blue{+7.4})                            \\
$\surd$ & $\surd$  & $\surd$ & \textbf{80.0}$\ $(\blue{+\textbf{8.8}}) & \textbf{82.3}$\ $(\blue{+\textbf{9.5}})                         \\\hline
\end{tabular}
\caption{Comparative results on CUHK-SYSU when employing different context learning strategies. }
\label{tab:cuhk1}
\end{table}

\textbf{Detection Context}. We visualize the contributions of the instance-to-instance term and instance-to-background term in Figure \ref{fig:contexts}(a) and \ref{fig:contexts}(b), by varying $\alpha_1$ and $\alpha_2$. We observe that both terms play positive roles in re-id feature learning, while the best performance is achieved with $\lambda_1 = 1$ and $\lambda_2 = 0.25$. %To differentiate the re-id features of pedestrians from the background, some prior works \cite{DBLP:conf/cvpr/ZhengZSCYT17,DBLP:conf/cvpr/ChenZYS20} employ confidence weighted similarity (CWS) for person matching, i.e., multiply the re-id features with the detection confidence. 
To differentiate the re-id features of pedestrians from the background, we employ the instance-to-background term to directly push the two types of features apart, which proves to be effective in our framework. 

\textbf{Memory Context}. To evaluate the effectiveness of the memory context, we compare it with the other widely employed ``sample by number" strategy, i.e., selecting the top $\lambda N^{-}$ hardest negatives from all the negative samples. As can be observed from Figure \ref{fig:contexts}(c), our model is sensitive to this hyperparameter. We observe that when $\lambda < 0.4$, the performance significantly decreases as too many easy samples are ignored. However, by selecting a proper hard negative ratio (i.e., 0.4 $\sim$ 0.8), our proposed strategy achieves notable improvements compared with the ``sample by number" strategy, as well as the baseline method without hard negative sampling ($\lambda = 1$). These results validate the effectiveness of the hard negatives sampled by our strategy.

\textbf{Scene Context}. To reveal how the scene context impacts the clustering results, we visualize the evolution of cluster numbers during training. As shown in Figure \ref{fig:contexts}(d), the number of clusters decreases smoothly, indicating that more and more samples are clustered together. However, there exists a significant gap with and without the scene context constraint, i.e., a large number of (about 10,000) samples are incorrectly clustered at each epoch. We employ the scene context to remove these incompatible samples from the clusters, hence improving the quality of the pseudo-labels. As scene context directly affects the pseudo-labels, we notice that it contributes the most among all the three context clues, demonstrating its importance.

\subsection{Comparison with Supervised Models}

\begin{table}[t]
\footnotesize
% \scriptsize
\setlength{\abovecaptionskip}{2mm}
\centering
\begin{tabular}{p{0.2cm}|p{3.4cm}|p{0.35cm}<{\centering}p{0.35cm}<{\centering}|p{0.35cm}<{\centering}p{0.35cm}<{\centering}}
\hline\thickhline
\rowcolor{mygray} 
\multicolumn{2}{c|}{}  & \multicolumn{2}{c|}{CUHK}                       & \multicolumn{2}{c}{PRW}                             \\ \cline{3-6} 
\rowcolor{mygray} 
\multicolumn{2}{c|}{\multirow{-2}{*}{Methods}}                       & \multicolumn{1}{c}{mAP} & \multicolumn{1}{c|}{top-1} & \multicolumn{1}{c}{mAP} & \multicolumn{1}{c}{top-1} \\ \hline\hline
\multirow{16}{*}{ \rotatebox{90}{fully supervised}}               & OIM~\cite{DBLP:conf/cvpr/XiaoLWLW17}      & 75.5  & 78.7      & 21.3   & 49.4    \\ 
& IDE~\cite{DBLP:conf/cvpr/ZhengZSCYT17}        & -  & - & 20.5   & 48.3 \\
 & IAN~\cite{DBLP:journals/pr/XiaoXTHWF19}    & 76.3  & 80.1 & 23.0   & 61.9 \\ 
 & NPSM~\cite{DBLP:conf/iccv/LiuFJKZQJY17}        & 77.9  & 81.2 & 24.2   & 53.1 \\
 & RCAA~\cite{DBLP:conf/eccv/ChangHSLYH18}        & 79.3  & 81.3 & -   & - \\
  & MGTS~\cite{DBLP:conf/eccv/ChenZOYT18}        & 83.0  & 83.7 & 32.6   & 72.1 \\
 & CTXG~\cite{DBLP:conf/cvpr/YanZNZXY19} & 84.1  & 86.5 & 33.4   & 73.6 \\
%   & CLSA~\cite{DBLP:conf/eccv/LanZG18}        & 87.2  & 88.5 & 38.7   & 65.0 \\
%  & QEEPS~\cite{DBLP:conf/cvpr/MunjalATG19} & 88.9  & 89.1 & 37.1   & 76.7 \\
 & BINet~\cite{DBLP:conf/cvpr/DongZST20a}        & 90.0  & 90.7 & 45.3   & 81.7 \\
%  & NAE~\cite{DBLP:conf/cvpr/ChenZYS20}        & 91.5 & 92.4 & 43.3   & 80.9 \\
 & NAE+~\cite{DBLP:conf/cvpr/ChenZYS20}        & 92.1 & 92.9 & 44.0   & 81.1 \\
 & IGPN~\cite{DBLP:conf/cvpr/DongZST20}        & 90.3  & 91.4 & 47.2   & 87.0 \\
 & RDLR~\cite{DBLP:conf/iccv/HanYZTZGS19}        & 93.0 & 94.2 & 42.9  & 70.2 \\
  & TCTS~\cite{DBLP:conf/cvpr/WangMCSC20}       & 93.9  & 95.1 & 46.8   & 87.5 \\
 & AlignPS~\cite{Yan_2021_CVPR} &93.1 & 93.4 &45.9 & 81.9 \\
 & SeqNet\cite{DBLP:conf/aaai/LiM21}        & \textbf{94.8} & \textbf{95.7} & \textbf{47.6}   & \textbf{87.6} \\
%  & \textbf{Our Baseline} &\textbf{81.1} & \textbf{80.6} &\textbf{26.3} & \textbf{59.6} \\
 
 \hline \hline
 \multicolumn{2}{l|}{\textbf{Ours} (weakly sup)}  
  &\textbf{80.0} & \textbf{82.3} &\textbf{16.2} & \textbf{68.0} \\

  \hline

\end{tabular}
\caption{Comparison with the state-of-the-art supervised person search models.}
\label{tab:sota}
\end{table}

% \begin{figure}[t]
% \setlength{\abovecaptionskip}{2mm}
% \begin{center}
%     %\fbox{\rule{0pt}{2in} %\rule{0.9\linewidth}{0pt}}
%     \includegraphics[width=\linewidth]{fig/sup.pdf}
% \end{center}
% \vspace{-12pt}
%   \caption{Comparison between our weakly supervised model and the supervised models on CUHK-SYSU. }
% \label{fig:cmp}
% \end{figure}

To further evaluate the performance of our weakly supervised framework, we compare it with the fully supervised models. Surprisingly, our weakly supervised framework outperforms several supervised models on CUHK-SYSU, e.g., OIM~\cite{DBLP:conf/cvpr/XiaoLWLW17}, IAN~\cite{DBLP:journals/pr/XiaoXTHWF19} and NPSM~\cite{DBLP:conf/iccv/LiuFJKZQJY17}. Compared with the state-of-the-art supervised models, e.g., SeqNet~\cite{DBLP:conf/aaai/LiM21}, NAE~\cite{DBLP:conf/cvpr/ChenZYS20}, BINet~\cite{DBLP:conf/cvpr/DongZST20a}, there still exists a significant margin in performance. We therefore hope this work will provide a starting point to enable future works to explore solutions for bridging this gap.

\begin{table}[t]
% \footnotesize
\setlength{\abovecaptionskip}{2mm}
\centering
\begin{tabular}{p{3.0cm}p{1.1cm}<{\centering}p{0.6cm}<{\centering}p{0.8cm}<{\centering}p{0.8cm}<{\centering}}
\hline\thickhline
\rowcolor{mygray} 
Methods &GFLOPS  &  Time &  mAP &  top1 \\  \hline \hline     
FRCNN + SBL  & 346  & 101 & 67.2 & 68.9 \\
FRCNN + SPCL & 346  & 100 & 71.8 & 72.1\\
FRCNN + MMT  & 346  & 101 & 73.4  & 74.9\\
%FRCNN + SSL \cite{DBLP:conf/cvpr/LinXWY020} & -  & - & - & -\\
\textbf{Ours} & \textbf{281}  & \textbf{68} & \textbf{80.0} & \textbf{82.3}\\
\hline
\end{tabular}
\caption{Comparison with two-step models on CUHK-SYSU, w.r.t. performance and efficiency. SBL denotes the strong baseline in \cite{DBLP:conf/nips/Ge0C0L20}. Runtime is measured by the average inference time in millisecond (ms).}
%``dconv'' stands for deformable conv.}
\label{tab:runtime}
%\vspace{-4mm}
\end{table}

\subsection{Comparison with Two-Step Models}
To evaluate the performance and efficiency of the proposed framework, we compare it with several two-step models, which first localize pedestrians with a detector (Faster-RCNN), and then employ an unsupervised re-id method \cite{DBLP:conf/nips/Ge0C0L20,DBLP:conf/iclr/GeCL20} for person retrieval. As can be seen from Table \ref{tab:runtime}, our one-step method not only outperforms the two-step models, but also displays significant advantage in terms of efficiency. Specifically, although our model is based on the re-id training prototype SPCL \cite{DBLP:conf/nips/Ge0C0L20}, the context information helps it outperform its two-step counterpart by a large margin. For the model complexity and runtime analysis, we employ the same backbone (i.e., ResNet-50) for all the two-step models. Therefore, they obtain similar FLOPS and runtime during inference. As our model only needs a single forward pass to generate both detection results and re-id features, it displays lower FLOPS (346G $\rightarrow$ 281G) and shorter runtime (101 ms $\rightarrow$ 68 ms). It is also noteworthy that the parameters of the two-step models are twice as our framework, as they adopt two separate ResNet-50 models for detection and re-id, respectively.

\subsection{Further Discussions}
% In this section, we provide further discussions on the factors that directly impact the performance of our weakly supervised person search model.

\begin{table}[t]
\setlength{\abovecaptionskip}{2mm}
\centering
\begin{tabular}{p{2cm}p{1.1cm}<{\centering}p{1.0cm}<{\centering}p{1.1cm}<{\centering}p{1.1cm}<{\centering}}
\hline\thickhline
\rowcolor{mygray} 
Methods &GFLOPS &Time  &  mAP &  top1 \\  \hline \hline     
ResNet-18 & \textbf{146} & \textbf{31}  & 60.1 & 63.2 \\
ResNet-34 & 235  & 40 & 67.5  & 70.8\\
ResNet-50 & 281 & 68  & 80.0  & 82.3\\
ResNet-101 & 327 & 81  & \textbf{80.4} & \textbf{82.5}\\
\hline
\end{tabular}
\caption{Results on CUHK-SYSU with various backbones.}
%``dconv'' stands for deformable conv.}
\label{tab:backbone}
%\vspace{-4mm}
\end{table}

\begin{figure*}[t]
\setlength{\abovecaptionskip}{0mm}
    \begin{center}
    %\fbox{\rule{0pt}{2in} %\rule{0.9\linewidth}{0pt}}
    \includegraphics[width=\linewidth]{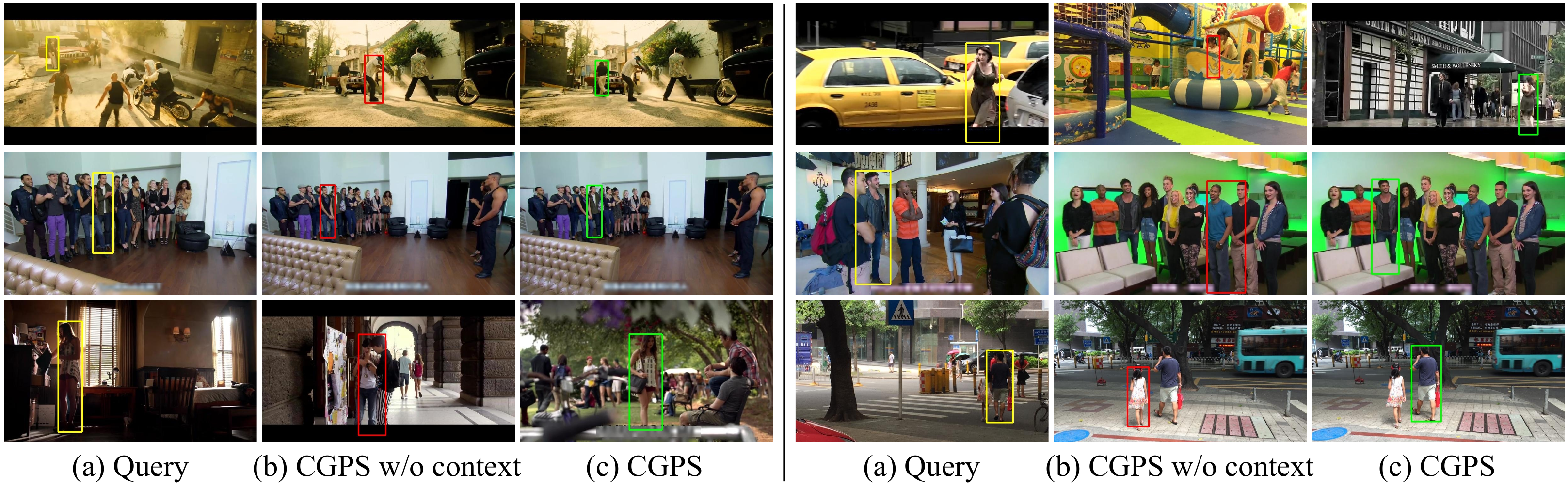}
\end{center}
\vspace{-2mm}
    \caption{Difficult cases that can be successfully retrieved by CGPS, but not CGPS without context information. The yellow bounding boxes denote the queries, while the green and red ones denote correct and incorrect top-1 matches, respectively.}
    \label{fig:qualitative}
\end{figure*}

\textbf{Different Backbones}. 
% We analyze the influence of different backbone networks in our weakly supervised setting, by employing different structures from the ResNet family.
As shown in Table \ref{tab:backbone}, using different backbones considerably affects our model's performance. With ResNet-18, the model achieves 60.1\% in mAP, while its performance is significantly improved to 80.4\% when employing ResNet-101. However, deeper backbone networks also bring significantly higher FLOPS (146G $\rightarrow$ 327G) and longer runtime (31 ms $\rightarrow$ 81 ms). In practice, a suitable backbone needs to be carefully selected to meet the requirements of real-world applications.

% \begin{table}[t]
% \setlength{\abovecaptionskip}{2mm}
% \centering
% \begin{tabular}{p{2cm}p{1.1cm}<{\centering}p{1.0cm}<{\centering}p{1.1cm}<{\centering}p{1.1cm}<{\centering}}
% \hline\thickhline
% \rowcolor{mygray} 
% Input Size &GFLOPS &Time  &  mAP &  top1 \\  \hline \hline     
% 800 $\times$ 480 & \textbf{259} & \textbf{66}  & 70.8 & 73.1 \\
% 1000 $\times$ 600 & 281  & 68 & 73.9  & 76.0\\
% 1200 $\times$ 720 & 307 & 76  & 74.1  & 75.7\\
% 1400 $\times$ 840 & 339 & 83  & 74.5 & 76.8\\
% 1600 $\times$ 960 & 374 & 88  & \textbf{75.0} & \textbf{77.1}\\
% \hline
% \end{tabular}
% \caption{Results on CUHK-SYSU with different input sizes.}
% %``dconv'' stands for deformable conv.}
% \label{tab:input}
% %\vspace{-4mm}
% \end{table}

% \textbf{Input Image Size}. As shown in Table \ref{tab:input}, we also find positive correlation between the model's performance and the input size. Specifically, enlarging the input size from $800 \times 480$ to $1000 \times 600$ leads to a performance gain of 3.1\% in mAP. However, continuing to increase the input size only yields marginal improvements, while introducing a large computational overhead. Therefore, the input size of $1000 \times 600$ achieves a good trade-off between performance and efficiency in our framework. 

\begin{figure}[t]
\setlength{\abovecaptionskip}{2mm}
    \centering
    %\fbox{\rule{0pt}{1in} 
    %\rule{0.9\linewidth}{0pt}}
    \vspace{-2mm}
    \includegraphics[width=\linewidth]{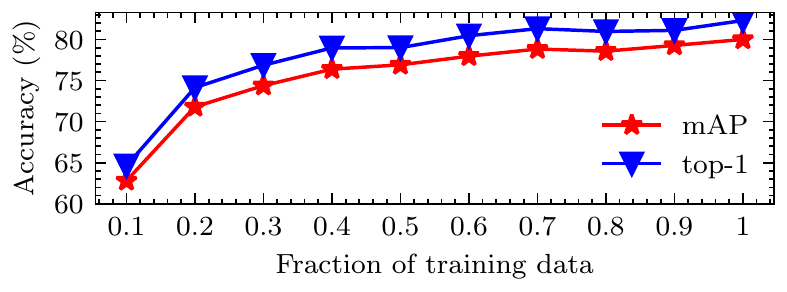}
    \caption{Results on CUHK-SYSU with different numbers of samples for weakly supervised training.}
    \label{fig:samples}
\end{figure}

\textbf{Training Samples}.
In Figure \ref{fig:samples}, we illustrate the impact of employing different numbers of training samples on CUHK-SYSU. We observe that more training samples generate better results, which motivates us to employ external training data. %Therefore, in the future, we can collect more weakly annotated data to further enhance the performance of our model. 
Following this idea, we combine CUHK-SYSU, PRW, and an external dataset COCO \cite{DBLP:conf/eccv/LinMBHPRDZ14} for training. Specifically, we select 6,500 images containing pedestrians from the original COCO dataset, and only utilize pedestrian bounding box annotations. As shown in Table \ref{tab:train1}, the combination of CUHK-SYSU and PRW achieves the best performance on both datasets. Furthermore, the improvement on PRW is more significant, maybe due to the fact that PRW contains fewer training samples. However, adding more training data does not always bring improvement, e.g., the combination of CUHK-SYSU and COCO achieves inferior performance compared with employing CUHK-SYSU alone. This is because of the domain gaps between these datasets; therefore, considering domain adaptation is another direction worth exploration in the future.

\begin{table}[t]
\setlength{\abovecaptionskip}{2mm}
\centering
\begin{tabular}{p{0.8cm}<{\centering}p{0.7cm}<{\centering}p{0.8cm}<{\centering}|p{0.8cm}<{\centering}p{0.8cm}<{\centering}|p{0.8cm}<{\centering}p{0.8cm}<{\centering}}
\hline\thickhline
\rowcolor{mygray}  
\multicolumn{3}{c|}{Training Data} 
            & \multicolumn{2}{c|}{CUHK-SYSU}    & \multicolumn{2}{c}{PRW}                     \\ \cline{1-7} 
\rowcolor{mygray}  
CUHK              & PRW               & COCO                & mAP                  & \multicolumn{1}{c|}{top-1} & mAP                  & \multicolumn{1}{c}{top-1}\\ 
\hline \hline  
$\surd$ &   & & 80.0 & 82.3          & 15.0 & 67.6     \\
 & $\surd$  & & 60.5 & 63.1    & 16.2 & 68.0         \\
 $\surd$   & $\surd$   & & \textbf{80.5} & \textbf{82.8}    & \textbf{18.7} & \textbf{73.2}    \\
  $\surd$  &   & $\surd$ & 74.8 & 77.0    & 13.0 & 66.5   \\
  &    $\surd$   & $\surd$ & 71.7  & 74.3    & 16.4 & 69.7   \\
  $\surd$ &    $\surd$   & $\surd$ & 75.8	 & 78.6    & 15.7	& 70.8  \\\hline
\end{tabular}
\caption{Comparative results on CUHK-SYSU and PRW when employing different combinations of training data.}
\label{tab:train1}
\end{table}

% \textbf{Training Data Combinations}
% To further reveal the impacts of combining different training data, we employ different configurations of CUHK-SYSU, PRW, and an external dataset COCO \cite{DBLP:conf/eccv/LinMBHPRDZ14} for training. Specifically, we select 6,500 images containing pedestrians from the original COCO dataset, and only utilize pedestrian bounding box annotations. As shown in Table \ref{tab:train1}, the combination of CUHK-SYSU and PRW achieves the best performance on both datasets. Furthermore, the improvement on PRW is more significant, maybe due to the fact that PRW contains fewer training samples. However, adding more training data does not always bring improvement, e.g., the combination of CUHK-SYSU and COCO achieves inferior performance compared with employing CUHK-SYSU alone. This is because of the domain gaps between these datasets; therefore, considering domain adaptation is another direction worth exploration in the future.

\begin{table}[t]
\setlength{\abovecaptionskip}{2mm}
\centering
\begin{tabular}{p{2.6cm}<{\centering}|p{0.9cm}<{\centering}p{0.9cm}<{\centering}|p{0.9cm}<{\centering}p{0.9cm}<{\centering}}
\hline\thickhline
\rowcolor{mygray}  
\multicolumn{1}{c|}{} 
            & \multicolumn{2}{c|}{CUHK-SYSU}    & \multicolumn{2}{c}{PRW}                     \\ \cline{2-5} 
\rowcolor{mygray}  
\multicolumn{1}{c|}{\multirow{-2}{*}{Annotation}}               & mAP                  & \multicolumn{1}{c|}{top-1} & mAP                  & \multicolumn{1}{c}{top-1}\\ 
\hline \hline  
Cascade R-CNN & 76.2 & 79.2           & 14.7 & 66.5          \\
RetinaNet  & 76.0 & 78.5           & 15.1 & 65.8          \\
EMD+RM  & 76.6 & 80.1           & 14.6 & 66.6          \\
GT & 80.0 & 82.3           & 16.2 & 68.0           \\
\hline
\end{tabular}
\caption{Comparative results on CUHK-SYSU and PRW when employing different bounding box annotations. }
\label{tab:data2}
%\vspace{-4mm}
\end{table}

\textbf{Extension to Unsupervised Person Search}.
Our weakly supervised framework can be naturally extended to a fully unsupervised setting by generating the bounding box annotations from existing pedestrian detectors, e.g, Cascade R-CNN \cite{DBLP:conf/cvpr/CaiV18}, RetinaNet \cite{DBLP:conf/iccv/LinGGHD17}, and EMD+RM \cite{DBLP:conf/cvpr/ChuZ0020}. We compare the model trained with the ground-truth (GT) bounding boxes and the predicted ones from these pedestrian detectors trained on \emph{CrowdHuman} \cite{DBLP:journals/corr/abs-1805-00123}. The comparison results are reported in Table \ref{tab:data2}. We observe that no matter which detector is employed, we can obtain satisfactory performance on both datasets, which is only slightly lower compared with the models trained with GT bounding boxes. This shows the potential of building a person search framework completely free of manual annotations.

\textbf{Qualitative Results}.
We visualize some qualitative results in Figure \ref{fig:qualitative}. As can be observed, our model successfully retrieves the target person in several challenging cases, where the baseline model without the  context information fails. As can be seen from the top-left example, CGPS retrieves the correct target from a different viewpoint. In contrast, the baseline model returns another person with similar appearance. Other examples also show that our context-guided model is more robust to the situations of occlusion, as well as illumination/scale variations. These results demonstrate the importance of the context clues in our framework. 

\section{Conclusion}
This paper investigates weakly supervised person search, a novel task with only bounding box locations, avoiding the need to collect labor-intensive identity annotations.
By extensively exploring the detection, memory, and scene context, we successfully develop a weakly supervised person search framework. Our framework achieves promising results on two benchmarks, but there still exists a gap compared with supervised models. We expect that this work will foster future research towards solving this.  

\bibliography{aaai22}

\end{document}